\documentclass{article}

     \usepackage[final]{cml4impact_2022}

\usepackage[utf8]{inputenc} 
\usepackage[T1]{fontenc}    
\usepackage{hyperref}       
\usepackage{url}            
\usepackage{booktabs}       
\usepackage{amsfonts}       
\usepackage{nicefrac}       
\usepackage{xcolor}         

\usepackage{bm}
\usepackage{amsmath}
\usepackage{amsthm}
\usepackage[english]{babel}
\newtheorem{theorem}{Theorem}

\usepackage{graphicx}
\usepackage{arydshln}
\usepackage{caption}
\usepackage{subcaption}

\usepackage{tikz}
\usetikzlibrary{fit,positioning,backgrounds}

\setcitestyle{square}
\setcitestyle{numbers}

\newcommand{\customfootnotetext}[2]{{
\renewcommand{\thefootnote}{#1}
\footnotetext[0]{#2}}}

\title{Variational Causal Inference}

\author{
  Yulun Wu \\
  University of California, Berkeley \\
  \texttt{yulun\_wu@berkeley.edu} \\
  \And
  Layne C. Price, Zichen Wang, Vassilis N. Ioannidis, Robert A. Barton \& George Karypis \\
  Amazon \\
  \texttt{\{prilayne,zichewan,ivasilei,rab,gkarypis\}@amazon.com} \\
}

\begin{document}

\maketitle

\begin{abstract}
  Estimating an individual's potential outcomes under counterfactual treatments is a challenging task for traditional causal inference and supervised learning approaches when the outcome is high-dimensional (e.g. gene expressions, impulse responses, human faces) and covariates are relatively limited. In this case, to construct one's outcome under a counterfactual treatment, it is crucial to leverage individual information contained in its observed factual outcome on top of the covariates. We propose a deep variational Bayesian framework that rigorously integrates two main sources of information for outcome construction under a counterfactual treatment: one source is the individual features embedded in the high-dimensional factual outcome; the other source is the response distribution of similar subjects (subjects with the same covariates) that factually received this treatment of interest.
\end{abstract}

\section{Introduction}
\label{intro}

In traditional causal inference, individualized treatment effect (ITE) is typically estimated by fitting supervised learning models on the covariate-specific efficacy $p(Y | X, T)$ or $\mathbb E \left[Y | X, T\right]$ with observed outcome $Y$, covariates $X$ and treatment $T$. However, in cases such as the single-cell genetic perturbation datasets \cite{dixit2016perturb, norman2019exploring, schmidt2022crispr} where $Y$ has thousands of dimensions while $X$ consists of only a few categorical features, such model could hardly be relied on to produce useful individualized results. To construct outcome $Y'$ under a counterfactual treatment $T'$, it is important and necessary to learn the individual-specific efficacy $p(Y' | Y, X, T, T')$, which leverages the rich information embedded in the factual outcome $Y$ that cannot be recovered by the handful of covariates.

For example, given a cell with type A549 ($X$) that received SAHA drug treatment ($T$), we may want to know what its gene expression profile ($Y'$) would look like if he had received Dex drug treatment ($T'$) instead. In this case, we would want to take the profiles of other cells with type A549 that have indeed received Dex drug treatment as reference, but would also want to extract as much individual features as possible from this cell's own expression profile ($Y$) to combine with the reference so that the reconstructed counterfactual expression profile could preserve its own identity.

Yet the lack of observability of $Y'$ presents a major difficulty for supervised learning when $Y$ is taken as an input. In previous works that involve self-supervised counterfactual generators \cite{louizos2017causal, yoon2018ganite, lotfollahi2021learning}, there is no explicit regulation on the trade-off between covariate-specific efficacy (reference) and residual individual features (identity). In \citet{louizos2017causal} and \citet{yoon2018ganite}, there is no such regulation; the trade-off is black-boxed and has to depend on network sizing. In \citet{lotfollahi2021learning}, the trade-off is implicitly encouraged by the balancing between adversarial reward and reconstruction loss. In this work, we present a semi-autoencoding framework with variational Bayes \citep{kingma2013auto, rezende2014stochastic} that is able to encode latents that explicitly honor covariate-specific efficacy while preserving individuality.

\section{Proposed Method}

\subsection{Semi-autoencoding Potential Outcomes}
\label{semi-autoencode}

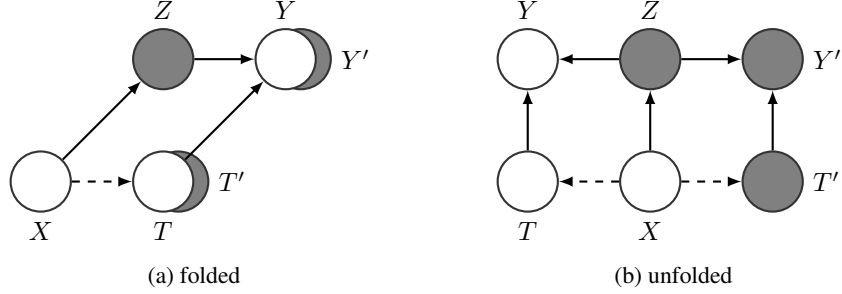
\begin{figure}
    \centering
    \begin{subfigure}[b]{0.45\textwidth}
        \centering
        \begin{tikzpicture}
            \tikzstyle{main}=[circle, minimum size = 8mm, thick, draw =black!80, node distance = 8mm]
            \tikzstyle{connect}=[-latex, thick]
            \tikzstyle{box}=[rectangle, draw=black!100]
              \node[main, fill = white!100] (X) [label=below:$X$] { };
              \node[main, fill = white!100] (T) [right=of X,label=below:$T$] { };
              \begin{scope}[on background layer]
                \node[main, fill = black!50] (tp) [right=of X,label=right:$T'$, xshift=2mm] { };
              \end{scope}
              \node[main, fill = black!50] (Z) [above=of T,label=above:$Z$] {};
              \node[main, fill = white!100] (Y) [right=of Z,label=above:$Y$] { };
              \begin{scope}[on background layer]
                \node[main, fill = black!50] (Yp) [right=of Z,label=right:$Y'$, xshift=2mm] { };
              \end{scope}
              \path (X) edge [connect] (Z)
                    (X) edge [dashed,connect] (T)
            		(Z) edge [connect] (Y)
            		(T) edge [connect] (Y);
        \end{tikzpicture}
        \caption{folded}
        \label{causal_diagram-folded}
    \end{subfigure}
    \begin{subfigure}[b]{0.45\textwidth}
        \centering
        \begin{tikzpicture}
            \tikzstyle{main}=[circle, minimum size = 8mm, thick, draw =black!80, node distance = 8mm]
            \tikzstyle{connect}=[-latex, thick]
            \tikzstyle{box}=[rectangle, draw=black!100]
              \node[main, fill = white!100] (X) [label=below:$X$] { };
              \node[main, fill = black!50] (Tp) [right=of X,label=right:$T'$] { };
              \node[main, fill = white!100] (T) [left=of X,label=below:$T$] { };
              \node[main, fill = black!50] (Z) [above=of X,label=above:$Z$] {};
              \node[main, fill = black!50] (Yp) [above=of Tp,label=right:$Y'$] { };
              \node[main, fill = white!100] (Y) [above=of T,label=above:$Y$] { };
              \path (X) edge [connect] (Z)
                    (X) edge [dashed,connect] (T)
                    (X) edge [dashed,connect] (Tp)
            		(Z) edge [connect] (Y)
            		(Z) edge [connect] (Yp)
            		(T) edge [connect] (Y)
            		(Tp) edge [connect] (Yp);
        \end{tikzpicture}
        \caption{unfolded}
        \label{causal_diagram-unfolded}
    \end{subfigure}
    \caption{The causal relation diagram. Each individual has a covariate-dependant feature state $Z$. Treatment $T$ (or counterfactual treatment $T'$) along with $Z$ determines outcome $Y$ (or counterfactual outcome $Y'$). In the causal diagram, white nodes are observed and dark grey nodes are unobserved; dashed edges exist if the data were not generated from a completely randomized trial.}
    \label{causal_diagram}
\end{figure}

\begin{figure}
    \centering
    \begin{subfigure}[b]{0.45\textwidth}
        \centering
        \begin{tikzpicture}
            \tikzstyle{main}=[circle, minimum size = 8mm, thick, draw =black!80, node distance = 8mm]
            \tikzstyle{connect}=[-latex, thick]
            \tikzstyle{box}=[rectangle, draw=black!100]
              \node[main, fill = white!100] (X) [label=below:$X$] { };
              \node[main, fill = white!100] (T) [right=of X,label=below:$T$] { };
              \node[main, fill = black!50] (Z) [above=of X,label=above:$Z$] {};
              \node[main, fill = white!100] (Y) [left=of X,label=below:$Y$] { };
              \path (X) edge [connect] (Z)
                    (Y) edge [connect] (Z)
            		(T) edge [connect] (Z);
        \end{tikzpicture}
        \caption{encoder}
        \label{encoder}
    \end{subfigure}
    \begin{subfigure}[b]{0.45\textwidth}
        \centering
        \begin{tikzpicture}
            \tikzstyle{main}=[circle, minimum size = 8mm, thick, draw =black!80, node distance = 8mm]
            \tikzstyle{connect}=[-latex, thick]
            \tikzstyle{box}=[rectangle, draw=black!100]
              \node[main, fill = black!50] (Z) [label=below:$Z$] { };
              \node[main, fill = white!100] (T) [right=of Z,label=below:$T$] { };
              \begin{scope}[on background layer]
                \node[main, fill = black!10] (Tp) [right=of Z,label=right:$T'$, xshift=2mm] { };
              \end{scope}
              \node[main, fill = white!100] (Y) [above=of T,label=above:$Y$] { };
              \begin{scope}[on background layer]
                \node[main, fill = black!50] (Yp) [above=of T,label=right:$Y'$, xshift=2mm] { };
              \end{scope}
              \node[main, fill = white!100] (X) [right=of T,label=below:$X$] { };
              \path (Z) edge [connect] (Y)
                    (T) edge [connect] (Y)
                    (X) edge [dashed,connect, bend left] (Tp);
        \end{tikzpicture}
        \caption{decoder}
        \label{decoder}
    \end{subfigure}
    \caption{Dependency structure of the encoder and decoder. White nodes are observed, light grey node is assigned (sampled) and dark grey nodes are inferred; dashed edge is optional. Note that the decoder estimates the conditional outcome distribution of $Y'$, in which case $T'$ need not necessarily be sampled according to a certain true distribution $p(T' | X)$ during optimization.}
    \label{semi-autoencoder}
\end{figure}
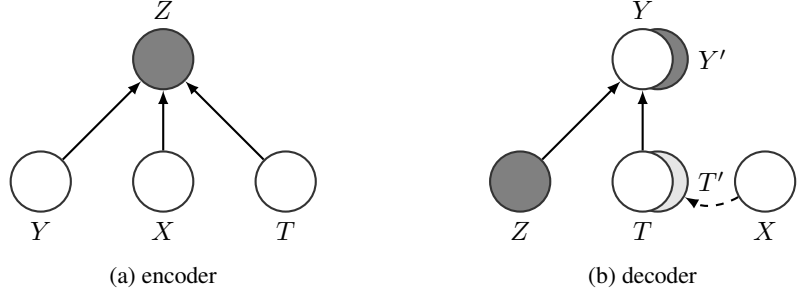

Let outcome $Y: \Omega \rightarrow (\mathcal{Y}, \Sigma_\mathcal{Y})$ be a random vector, $X: \Omega \rightarrow (\mathcal{X}, \Sigma_\mathcal{X})$ be a mix of categorical and real-valued covariates, $T: \Omega \rightarrow (\mathcal{T}, \Sigma_\mathcal{T})$ be a categorical or real-valued treatment (or multiple treatments) and $Z: \Omega \rightarrow \mathbb{R}^d$ be a real-valued latent feature vector on a probability space $(\Omega, \Sigma, P)$. Suppose the causal relations between random variables (and random vectors) follow a structural causal model (SCM) \cite{pearl1995causal} depicted by the causal diagram in Figure \ref{causal_diagram}, where we formulate counterfactuals $Y'$ and $T'$ as separate variables apart from $Y$ and $T$, but having a conditional outcome distribution $p(Y' | Z, T'=a)$ identical to that of its factual counterpart $p(Y | Z, T=a)$ on a treatment level $a$. $Y'$ and $T'$ are parts of the full random variable collection $W=(X, Z, T, T', Y, Y')$, yet are never observed. Latent $Z$ is designed to be a feature vector of $Y$ such that random factor $U_Y$ in structural equation $Y=f(Z, T, U_Y)$ is significantly less noisy than that of the outcome equation $Y=f(X, T, U_Y)$ in a traditional SCM of triplets $(Y, X, T)$. Under this setting, $Z$ has a posterior distribution $p(Z | Y, X, T)$ given the observations; $Y$ and $Y'$ can be reconstructed by $p(Y | Z, T)$ and $p(Y' | Z, T')$ respectively if $Z$ is recovered. Similar to prior works on autoencoder \cite{vincent2008extracting, bengio2014deep}, we estimate the latent recognition model and outcome reconstruction model with deep neural networks $q_\phi$ and $p_\theta$, whose dimensionalities control the noise level of $U_Y$. Figure \ref{semi-autoencoder} shows the graphical model for the encoder and decoder.

While the reconstruction of $Y$ is self-supervised, we can only assess the construction of $Y'$ by looking at its resemblance to similar individuals that indeed received treatment $T'$. Hence we may develop a semi-autoencoding scheme that works as follows. During training, we estimate $Z$ given the observed triplets $(Y, X, T)$, then predict both factual and counterfactual outcomes $\tilde{Y}_{\theta,\phi}$ and $\tilde{Y}'_{\theta,\phi}$. Inputting $T$ into decoder $p_\theta$ yields $\tilde{Y}_{\theta,\phi}$, which is then evaluated by the reconstruction loss $L(\tilde{Y}_{\theta,\phi}, Y)$; inputting $T'$ into $p_\theta$ yields $\tilde{Y}'_{\theta,\phi}$, which we evaluate by the negative likelihood loss $-\mathcal{L}_{p(Y' | X, T')}(\tilde{Y}'_{\theta,\phi})$ with respect to the counterfactual outcome distribution $p(Y' | X, T')$. The intuition is that, if $\tilde{Y}'_{\theta,\phi}$ is indeed one's outcome under $T'$, then the likelihood of $\tilde{Y}'_{\theta,\phi}$ coming from the outcome distribution of individuals with the same attributes that factually received treatment $T'$  is supposed to be high. The total loss of the semi-autoencoder can then be formed as a weighted combination:
\begin{align}
\label{sae_loss}
    L(\theta, \phi) = L(\tilde{Y}_{\theta,\phi}, Y) -\omega\cdot \mathcal{L}_{\hat{p}(Y' | X, T')}(\tilde{Y}'_{\theta,\phi})
\end{align}
where $\omega$ is a scaling coefficient and $\hat{p}$ is the traditional covariate-specific outcome model fit on the observed variables $(X, T, Y)$ (notice that $p(Y' | X, T'=a)=p(Y | X, T=a)$ for any $a$). If $\hat{p}$ is difficult to fit, we can train a discriminator $\mathcal{D}(X, T, Y)$ for $p(Y | X, T)$ (and $p(Y' | X, T')$) instead with the adversarial approach \cite{goodfellow2014generative} and use its discrimination loss over counterfactuals $L_\mathcal{D}(X, T', \tilde{Y}'_{\theta,\phi})$ in place of $-\mathcal{L}_{\hat{p}(Y' | X, T')}(\tilde{Y}'_{\theta,\phi})$. In cases where covariates are limited and discrete such as the single-cell perturbation datasets, $\hat{p}$ can simply be the smoothened empirical outcome distribution under treatment $T$ stratified by covariates $X$.

\subsection{Variational Causal Inference}

Now we present our main result by rigorously formulating the objective, providing a probabilistic theoretical backing, and specifying an optimization scheme that reflects the intuition described in the previous section. Suppose we want to optimize $p(Y' | Y, X, T, T')$ instead of the traditional outcome model $p(Y | X, T)$. The following theorem states the evidence lower bound (ELBO) and thus provides a roadmap for stochastic optimization:

\begin{theorem}
\label{elbo}
Suppose random vector $W=(X, Z, T, T', Y, Y')$ follows a causal structure defined by the Bayesian network in Figure \ref{causal_diagram}. Then $\log \left[ p (Y' | Y, X, T, T') \right]$ has the following variational lower bound:
\begin{align}
    \log \left[ p (Y' | Y, X, T, T') \right] &\geq \mathbb E_{p (Z | Y, X, T)} \log \left[ p (Y | Z, T) \right] - D \left[ p (Y | X, T) \parallel p (Y' | X, T') \right] \nonumber \\
    &\quad - D_\mathrm{KL} \left[ p (Z | Y, X, T) \parallel p (Z | Y', X, T') \right]
\end{align}
where $D [ p \parallel q ] = \log p - \log q$.
\end{theorem}

Proof of the theorem can be found in Appendix \ref{elbo_proof}. Theorem \ref{elbo} states that, in order to maximize $\log \left[ p (Y' | Y, X, T, T') \right]$, we can instead estimate and maximize the ELBO which consists of the expected reconstruction likelihood $E_{p (Z | Y, X, T)} \log \left[ p (Y | Z, T) \right]$ and covariate-specific outcome likelihood $p (Y' | X, T')$ of the counterfactuals w.r.t. $p (Y | X, T)$ of the factuals, echoing the intuition highlighted by Equation \ref{sae_loss}, with an additional divergence term $-D_\mathrm{KL} \left[ p (Z | Y, X, T) \parallel p (Z | Y', X, T') \right]$ that regularizes the similarity across latent distributions under different potential scenarios. 
The objective after weighting and parameterizing with deep neural networks is given as
\begin{align}
    \label{VCI-objective}
    J(\theta, \phi) &= \mathbb E_{q_\phi (Z | Y, X, T)} \log \left[ p_\theta (Y | Z, T) \right] + \omega_1 \cdot \log \left[ p (\tilde{Y}'_{\theta,\phi} | X, T') \right] \nonumber \\
    &\quad - \omega_2 \cdot D_\mathrm{KL} \left[ q_\phi (Z | Y, X, T) \parallel q_\phi (Z | \tilde{Y}'_{\theta,\phi}, X, T') \right]
\end{align}
where $\omega_1$ and $\omega_2$ are scaling coefficients; $\tilde{Y}'_{\theta,\phi} \sim E_{q_\phi (Z | Y, X, T)} p_\theta (Y' | Z, T')$ and $p (\tilde{Y}'_{\theta,\phi} | X, T')$ can be estimated by $\hat{p} (\tilde{Y}'_{\theta,\phi} | X, T')$ with the approaches described at the end of Section \ref{semi-autoencode}. $\hat{p}(Y | X, T)$ does not impose gradient on $(p_\theta, q_\phi)$ and is thus omitted in $J(\theta, \phi)$. Note that this framework is fundamentally different from variational autoencoder (VAE) \citep{kingma2013auto} based frameworks such as CVAE \cite{sohn2015learning} and CEVAE \cite{louizos2017causal} since the variational lower bound is derived directly from the causal objective and the KL-divergence term serves a causal purpose that is completely different from the latent bounding purpose of VAEs. An explanation of this divergence term is given below.

\paragraph{Divergence Interpretation} The divergence term $-D_\mathrm{KL} \left[ q_\phi (Z | Y, X, T) \parallel q_\phi (Z | \tilde{Y}'_{\theta,\phi}, X, T') \right]$ in Equation \ref{VCI-objective} encourages the preservation of individuality in counterfactual outcome constructions. In the autoencoding process, the latent variable $Z$ is required to preserve individual information contained in $Y$ beyond covariates $X$ in order to obtain a quality reconstruction of $Y$. Yet the counterfactual construction $p_\theta(Y' | \tilde{Z}_\phi, T')$ could discard this individuality in $Z$ once it realizes a counterfactual treatment $T'$ is in the inputs, if the outcome construction was only penalized by the covariate-specific likelihood loss $-\log \left[ \hat{p} (Y' | X, T') \right]$. Such behavior is regulated with the addition of the divergence term, since otherwise we would not be able to recover a latent distribution $q_\phi (Z | \tilde{Y}'_{\theta,\phi}, X, T')$ close to $q_\phi (Z | Y, X, T)$.

\subsection{Marginal Effect Estimation}

Upon obtaining individual level results, we introduce a robust estimator for the marginal treatment effect $\Psi(p) = \mathbb E_p [ Y'_{\mathrm{do}(T'=a)} ]$ of a treatment level $a$ in the variational causal inference (VCI) framework. By \citet{van2000asymptotic}, a sequence of estimators $\hat{\Psi}_n$ is asymptotically efficient if $\sqrt{n}(\hat{\Psi}_n - \Psi(p)) = 1 / \sqrt{n} \sum_{k=1}^n \tilde{\psi}_p(W_k) + o_p(1)$ where $\tilde{\psi}_p$ is the efficient influence function of $\Psi(p)$ and $W_k \sim p(W)$. The theorem below gives this efficient influence function and thus provides a construction of an asymptotically efficient regular estimator for $\Psi$:

\begin{theorem}
\label{variational-ATT}
Suppose $W: \Omega \rightarrow E$ follows a causal structure defined by the Bayesian network in Figure \ref{causal_diagram}, where the counterfactual conditional distribution $p(Y', T' | Z, X)$ is identical to that of its factual counterpart $p(Y, T | Z, X)$. Then $\Psi(p)$ has the following efficient influence function:
\begin{align}
    \tilde{\psi}_p(W) = \frac{I(T = a)}{p(T | X)} (Y - \mathbb E_p\left[ Y | Z, T \right]) + \mathbb E_p\left[ Y' | Z, T'=a \right] - \Psi.
\end{align}
\end{theorem}

Proof of the theorem can be found in Appendix \ref{variational-ATT_proof}. In practice, we only need $p(Y' | Z, T' = a)$ to be identical to its factual counterpart $p(Y | Z, T = a)$ given a treatment level $a$ such that they can be estimated with the same decoder $p_\theta$. Since $Y'$ and $T'$ are both unobserved in the data, it does not bother to simply acknowledge that $T'$ is sampled according to $p(T | X)$. By Theorem \ref{variational-ATT}, the following estimator is asymptotically efficient under some regularity conditions \cite{van2006targeted}:
\begin{align}
    \label{ATT-estimator}
    \hat{\Psi}_n(\theta, \phi) = \frac{1}{n} \sum_{k=1}^n \left\{ \frac{I(T_k = a)}{\hat{p}(T_k | X_k)} \cdot Y_k + \left(1-\frac{I(T_k = a)}{\hat{p}(T_k | X_k)}\right) \cdot \mathbb E_{p_\theta} \left[ Y' | \tilde{Z}_{k, \phi}, T_k'=a \right] \right\}
\end{align}
where $(Y_k, X_k, T_k)$ are the observed variables of the $k$-th individual and $\tilde{Z}_{k,\phi} \sim q_\phi (Y_k, X_k, T_k)$; $\hat{p}$ is an estimation of the propensity score. As can be seen, $\hat{\Psi}_n(\theta, \phi)$ takes a very similar form as the augmented inverse propensity weighted (AIPW) estimator of the average treatment effect on treated (ATT) in traditional causal inference.

\paragraph{Covariate-specific Marginal Effect} The robust estimator within the VCI framework can also be employed in estimating covariate-specific marginal treatment effect $\Xi(p) = \mathbb E_p [ Y'_{\mathrm{do}(X=c, T'=a)} ]$ for a given covariate $c$ of interest, by applying Equation \ref{ATT-estimator} on the set of observations with $X=c$. Note that the regression adjustment term $\mathbb E_{p_\theta} \left[ Y' | \tilde{Z}_{k, \phi}, T_k'=a \right]$ by our deep network design can vary across different subjects within the same covariate group, which is not true for that of the AIPW estimator in traditional causal inference.

\section{Experiments}
\label{experiments}

We tested our framework on single-cell perturbation datasets along with four benchmark models --- CEVAE \cite{louizos2017causal}, GANITE \cite{yoon2018ganite}, CPA \cite{lotfollahi2021learning} and a naive autoencoding method. In single-cell perturbation datasets, each observation is an $n$-dimensional vector (the outcome) storing the expression counts of $n$ selected genes in a single cell, along with the perturbation (the treatment) this cell was subjected to and the cell covariates. We employed the publicly available sci-Plex dataset from \citet{srivatsan2020massively} (Sciplex) and the CRISPRa dataset from \citet{schmidt2022crispr} (Marson) in our experiments. Two thousand most variable genes were selected from each dataset. Data with certain treatment-covariate combinations are held out as the out-of-distribution (OOD) set and the rest are split into training and validation set (four-to-one). Details of the hyperparameter settings can be found in Appendix \ref{hyperparam-setting}.

\paragraph{Out-of-Distribution Selections}
We randomly select a covariate category (e.g. a cell type) and hold out all cells in this category that received one of the twenty perturbations whose effects are the hardest to predict. We use these held-out data to compose the out-of-distribution (OOD) set. We computed the Euclidean distance between the pseudobulked gene expression of each perturbation against the rest of the dataset, and selected the twenty most distant ones as the hardest-to-predict perturbations. This is the same procedure carried out by \citet{lotfollahi2021learning}.

\paragraph{Differentially-Expressed Genes}
In order to evaluate the quality of the predictions on the genes that were substantially affected by the perturbations, we select sets of 50 differentially-expressed (DE) genes associated with each perturbation and separately evaluate model performance on these genes. This is the same procedure carried out by \citet{lotfollahi2021learning}.

\subsection{Out-of-Distribution Predictions}
\label{ood-pred}

Same as \citet{lotfollahi2021learning}, for each perturbation of each covariate level (e.g. each cell type of each donor) in the OOD set, we compute the $R^2$ (coefficient of determination) of the average outcome predictions for all genes and DE genes using samples from the validation set against the true empirical average over samples from the OOD set. The average $R^2$ over all perturbations of all covariate levels is then calculated as the evaluation metric. Table \ref{result-table1} shows the best results of each model over five independent runs.

\begin{table}[ht!] 
  \caption{Results on OOD sets}
  \label{result-table1}
  \centering
  \small
  \begin{tabular}{lcccc}
    \toprule
    & \multicolumn{2}{c}{Sciplex \cite{srivatsan2020massively}} & \multicolumn{2}{c}{Marson \cite{schmidt2022crispr}}\\
    \cmidrule(r){2-3} \cmidrule(r){4-5}
    & all genes & DE genes & all genes & DE genes \\
    \midrule
    AE\textsuperscript{$\S$}  & 0.785 & 0.443 & 0.824 & 0.458 \\
    CEVAE \cite{louizos2017causal}  & 0.781 & 0.448 & 0.814 & 0.437 \\
    GANITE\textsuperscript{$\P$} \cite{yoon2018ganite}  & 0.763 & 0.429 & 0.806 & 0.476 \\
    CPA \cite{lotfollahi2021learning}  & \bf 0.837 & 0.488 & 0.879 & 0.569 \\
    \hdashline
    VCI  & \bf 0.837 & \bf 0.504 & \bf 0.893 & \bf 0.658 \\
    \bottomrule
  \end{tabular}
\end{table}
\customfootnotetext{$\S$}{A naive approach adapting Autoencoder to counterfactual generation. See Appendix \ref{ae-adaptation}.}
\customfootnotetext{$\P$}{GANITE's counterfactual block. GANITE's counterfacutal generator does not scale with a combination of high-dimensional outcome and multi-level treatment, see Appendix \ref{ganite-adaptation} for the adaptation we made.}

In the experiments, we used the empirical outcome distribution (with Gaussian kernel smoother) stratified by $X$ and $T$ to estimate the covariate-specific outcome model $p(Y | X, T)$. There are a few gradient detaching options for $\tilde{Y}'_{\theta,\phi}$ when inputting $\tilde{Y}'_{\theta,\phi}$ into the KL-divergence term $D_\mathrm{KL} \left[ q_\phi (Z | Y, X, T) \parallel q_\phi (Z | \tilde{Y}'_{\theta,\phi}, X, T') \right]$: not detached; detached from encoder; detached from both encoder and decoder. Here we chose to not detach $\tilde{Y}'_{\theta,\phi}$. Our variational Bayesian causal inference framework beat all state-of-the-art models in these experiments, with the largest fractional improvement on DE genes which are most causally affected by the perturbations.

\subsection{Marginal Estimations}

In this section, we use the same evaluation metric, but compute each $R^2$ with the robust marginal estimator and compare the results to that of the empirical mean estimator. Note that on OOD set, the robust estimator reduces to empirical mean since no perturbation-covariates combination exist in validation set. Therefore, we compute the $R^2$ of the marginal estimators using samples from the training set against the true empirical average on the validation set in these experiments. In each run, we train a VCI model for individualized outcome predictions, and calculate the evaluation metric for each marginal estimator periodically during the course of training. Table \ref{result-table2} shows the results on Marson over five independent runs. Note that the goal of robust estimation is to produce less biased estimators with tighter confidence bounds, hence here we reported the mean and standard deviation across runs instead of the maximum to reflect its effectiveness regarding this goal.

\begin{table}[ht!] 
  \caption{Comparison of marginal estimators on Marson \cite{schmidt2022crispr}}
  \label{result-table2}
  \centering
  \small
  \begin{tabular}{lcccc}
    \toprule
    & \multicolumn{2}{c}{All Genes} & \multicolumn{2}{c}{DE Genes}\\
    \cmidrule(r){2-3} \cmidrule(r){4-5}
    Episode & mean & robust & mean & robust \\
    \midrule
    40  & 0.9141 $\pm$ 0.0159 & \textbf{0.9343} $\pm$ \textbf{0.0080} & 0.7108 $\pm$ 0.0735 & \textbf{0.9146} $\pm$ \textbf{0.0305} \\
    80  & 0.9171 $\pm$ 0.0104 & \textbf{0.9349} $\pm$ \textbf{0.0068} & 0.7274 $\pm$ 0.0462 & \textbf{0.9163} $\pm$ \textbf{0.0254} \\
    120 & 0.9204 $\pm$ 0.0097 & \textbf{0.9352} $\pm$ \textbf{0.0063} & 0.7526 $\pm$ 0.0447 & \textbf{0.9182} $\pm$ \textbf{0.0229} \\
    160 & 0.9157 $\pm$ \textbf{0.0043} & \textbf{0.9355} $\pm$ 0.0053 & 0.7383 $\pm$ 0.0402 & \textbf{0.9203} $\pm$ \textbf{0.0199} \\
    \bottomrule
  \end{tabular}
\end{table}

As is shown in the table, the robust estimator provides a crucial adjustment to the empirical mean of model predictions especially on the hard-to-predict elements of high-dimensional vectors. Such estimation could be valuable in many contexts involving high-dimensional predictions where deep learning models might plateau at rather low ceilings.

\section{Conclusion}
\label{conclusion}

In this work, we introduced a variational Bayesian causal inference framework for high-dimensional individualized treatment effect prediction. With this framework, covariate-specific efficacy and individual identity can be explicitly balanced and optimized. In experiments, our framework excelled the state-of-the-art algorithms at out-of-distribution predictions on single-cell perturbation datasets --- a widely accepted and biologically meaningful task. We note that although several approaches were described for the estimation of covariate-specific distribution loss, our most compelling method is for discrete treatments, where the distribution can be estimated empirically via stratification.

\begin{ack}
  We thank Ivana Malenica, Dominik Janzing, David E. Heckerman for the insightful discussions, and Carlo De Donno for the guidance on data processing.
\end{ack}

\bibliography{neurips_2022}

\begin{thebibliography}{18}
\providecommand{\natexlab}[1]{#1}
\providecommand{\url}[1]{\texttt{#1}}
\expandafter\ifx\csname urlstyle\endcsname\relax
  \providecommand{\doi}[1]{doi: #1}\else
  \providecommand{\doi}{doi: \begingroup \urlstyle{rm}\Url}\fi

\bibitem[Bengio et~al.(2014)Bengio, Laufer, Alain, and
  Yosinski]{bengio2014deep}
Yoshua Bengio, Eric Laufer, Guillaume Alain, and Jason Yosinski.
\newblock Deep generative stochastic networks trainable by backprop.
\newblock In \emph{International Conference on Machine Learning}, pp.\
  226--234. PMLR, 2014.

\bibitem[Dixit et~al.(2016)Dixit, Parnas, Li, Chen, Fulco, Jerby-Arnon,
  Marjanovic, Dionne, Burks, Raychowdhury, et~al.]{dixit2016perturb}
Atray Dixit, Oren Parnas, Biyu Li, Jenny Chen, Charles~P Fulco, Livnat
  Jerby-Arnon, Nemanja~D Marjanovic, Danielle Dionne, Tyler Burks, Raktima
  Raychowdhury, et~al.
\newblock Perturb-seq: dissecting molecular circuits with scalable single-cell
  rna profiling of pooled genetic screens.
\newblock \emph{cell}, 167\penalty0 (7):\penalty0 1853--1866, 2016.

\bibitem[Goodfellow et~al.(2014)Goodfellow, Pouget-Abadie, Mirza, Xu,
  Warde-Farley, Ozair, Courville, and Bengio]{goodfellow2014generative}
Ian Goodfellow, Jean Pouget-Abadie, Mehdi Mirza, Bing Xu, David Warde-Farley,
  Sherjil Ozair, Aaron Courville, and Yoshua Bengio.
\newblock Generative adversarial nets.
\newblock \emph{Advances in neural information processing systems}, 27, 2014.

\bibitem[Kingma \& Welling(2013)Kingma and Welling]{kingma2013auto}
Diederik~P Kingma and Max Welling.
\newblock Auto-encoding variational bayes.
\newblock \emph{arXiv preprint arXiv:1312.6114}, 2013.

\bibitem[Levy(2019)]{levy2019tutorial}
Jonathan Levy.
\newblock Tutorial: Deriving the efficient influence curve for large models.
\newblock \emph{arXiv preprint arXiv:1903.01706}, 2019.

\bibitem[Lotfollahi et~al.(2021)Lotfollahi, Susmelj, De~Donno, Ji, Ibarra,
  Wolf, Yakubova, Theis, and Lopez-Paz]{lotfollahi2021learning}
Mohammad Lotfollahi, Anna~Klimovskaia Susmelj, Carlo De~Donno, Yuge Ji,
  Ignacio~L Ibarra, F~Alexander Wolf, Nafissa Yakubova, Fabian~J Theis, and
  David Lopez-Paz.
\newblock Learning interpretable cellular responses to complex perturbations in
  high-throughput screens.
\newblock \emph{bioRxiv}, 2021.

\bibitem[Louizos et~al.(2017)Louizos, Shalit, Mooij, Sontag, Zemel, and
  Welling]{louizos2017causal}
Christos Louizos, Uri Shalit, Joris~M Mooij, David Sontag, Richard Zemel, and
  Max Welling.
\newblock Causal effect inference with deep latent-variable models.
\newblock \emph{Advances in neural information processing systems}, 30, 2017.

\bibitem[Norman et~al.(2019)Norman, Horlbeck, Replogle, Ge, Xu, Jost, Gilbert,
  and Weissman]{norman2019exploring}
Thomas~M Norman, Max~A Horlbeck, Joseph~M Replogle, Alex~Y Ge, Albert Xu, Marco
  Jost, Luke~A Gilbert, and Jonathan~S Weissman.
\newblock Exploring genetic interaction manifolds constructed from rich
  single-cell phenotypes.
\newblock \emph{Science}, 365\penalty0 (6455):\penalty0 786--793, 2019.

\bibitem[Pearl(1988)]{pearl1988probabilistic}
Judea Pearl.
\newblock \emph{Probabilistic reasoning in intelligent systems: networks of
  plausible inference}.
\newblock Morgan kaufmann, 1988.

\bibitem[Pearl(1995)]{pearl1995causal}
Judea Pearl.
\newblock Causal diagrams for empirical research.
\newblock \emph{Biometrika}, 82\penalty0 (4):\penalty0 669--688, 1995.

\bibitem[Rezende et~al.(2014)Rezende, Mohamed, and
  Wierstra]{rezende2014stochastic}
Danilo~Jimenez Rezende, Shakir Mohamed, and Daan Wierstra.
\newblock Stochastic backpropagation and approximate inference in deep
  generative models.
\newblock In \emph{International conference on machine learning}, pp.\
  1278--1286. PMLR, 2014.

\bibitem[Schmidt et~al.(2022)Schmidt, Steinhart, Layeghi, Freimer, Bueno,
  Nguyen, Blaeschke, Ye, and Marson]{schmidt2022crispr}
Ralf Schmidt, Zachary Steinhart, Madeline Layeghi, Jacob~W Freimer, Raymund
  Bueno, Vinh~Q Nguyen, Franziska Blaeschke, Chun~Jimmie Ye, and Alexander
  Marson.
\newblock Crispr activation and interference screens decode stimulation
  responses in primary human t cells.
\newblock \emph{Science}, 375\penalty0 (6580):\penalty0 eabj4008, 2022.

\bibitem[Sohn et~al.(2015)Sohn, Lee, and Yan]{sohn2015learning}
Kihyuk Sohn, Honglak Lee, and Xinchen Yan.
\newblock Learning structured output representation using deep conditional
  generative models.
\newblock \emph{Advances in neural information processing systems}, 28, 2015.

\bibitem[Srivatsan et~al.(2020)Srivatsan, McFaline-Figueroa, Ramani, Saunders,
  Cao, Packer, Pliner, Jackson, Daza, Christiansen,
  et~al.]{srivatsan2020massively}
Sanjay~R Srivatsan, Jos{\'e}~L McFaline-Figueroa, Vijay Ramani, Lauren
  Saunders, Junyue Cao, Jonathan Packer, Hannah~A Pliner, Dana~L Jackson,
  Riza~M Daza, Lena Christiansen, et~al.
\newblock Massively multiplex chemical transcriptomics at single-cell
  resolution.
\newblock \emph{Science}, 367\penalty0 (6473):\penalty0 45--51, 2020.

\bibitem[Van Der~Laan \& Rubin(2006)Van Der~Laan and Rubin]{van2006targeted}
Mark~J Van Der~Laan and Daniel Rubin.
\newblock Targeted maximum likelihood learning.
\newblock \emph{The international journal of biostatistics}, 2\penalty0 (1),
  2006.

\bibitem[Van~der Vaart(2000)]{van2000asymptotic}
Aad~W Van~der Vaart.
\newblock \emph{Asymptotic statistics}, volume~3.
\newblock Cambridge university press, 2000.

\bibitem[Vincent et~al.(2008)Vincent, Larochelle, Bengio, and
  Manzagol]{vincent2008extracting}
Pascal Vincent, Hugo Larochelle, Yoshua Bengio, and Pierre-Antoine Manzagol.
\newblock Extracting and composing robust features with denoising autoencoders.
\newblock In \emph{Proceedings of the 25th international conference on Machine
  learning}, pp.\  1096--1103, 2008.

\bibitem[Yoon et~al.(2018)Yoon, Jordon, and Van Der~Schaar]{yoon2018ganite}
Jinsung Yoon, James Jordon, and Mihaela Van Der~Schaar.
\newblock Ganite: Estimation of individualized treatment effects using
  generative adversarial nets.
\newblock In \emph{International Conference on Learning Representations}, 2018.

\end{thebibliography}
\bibliographystyle{neurips_2022}

\appendix
\section*{Appendix}

\section{Proof of Theorems}
\label{proofs}

\subsection{Proof of Theorem \ref{elbo}}
\label{elbo_proof}

\begin{proof}
    By the d-separation \cite{pearl1988probabilistic} of paths on the causal graph defined in Figure \ref{causal_diagram}, we have
    \begin{align}
        \log \left[ p (Y' | Y, X, T, T') \right]
     & = \log \mathbb E_{p (Z | Y, X, T)} \left[ 
        p (Y' | Z, Y, X, T, T') \right] \\
     & \geq \mathbb E_{p (Z | Y, X, T)} \log \left[ 
        p (Y' | Z, Y, X, T, T') \right] \quad \text{(Jensen's inequality)}\\
     & = \mathbb E_{p (Z | Y, X, T)} \log \frac{p (Y', Z | Y, X, T, T')}{p (Z | Y, X, T)} \\
     & = \mathbb E_{p (Z | Y, X, T)} \log \frac{p (Y', Z, Y | X, T, T')}{p (Z | Y, X, T) p (Y | X, T)} \\
     & = \mathbb E_{p (Z | Y, X, T)} \log \frac{p (Y | Z, T) p (Z | Y', X, T') p (Y' | X, T')}{p (Z | Y, X, T) p (Y | X, T)} \\
     & = \mathbb E_{p (Z | Y, X, T)} \log \left[ p (Y | Z, T) \right] - D_\mathrm{KL} \left[ p (Z | Y, X, T) \parallel p (Z | Y', X, T') \right] \nonumber \\
     &\quad - D \left[ p (Y | X, T) \parallel p (Y' | X, T') \right].
    \end{align}
    Reorganizing the terms yields the desired result.
\end{proof}


\subsection{Proof of Theorem \ref{variational-ATT}}
\label{variational-ATT_proof}

\begin{proof}
    $\Psi(p)$ has the identification $\Psi(p) = \mathbb E_p [\mathbb E_p [Y' | Z, X, T'=a ]] = \mathbb E_p [\mathbb E_p [Y' | Z, T'=a ]]$ under Figure \ref{causal_diagram}. Following \citet{van2000asymptotic}, we define a path $p_\epsilon(\Lambda)=p(\Lambda)(1+\epsilon S(\Lambda))$ on density $p$ of $\Lambda$ as a submodel that passes through $p$ at $\epsilon=0$ in the direction of the score $S(\Lambda)=\frac{d}{d\epsilon} \log \left[p_\epsilon(\Lambda)\right] \Big{\rvert}_{\epsilon=0}$. Following the key identity presented in \citet{levy2019tutorial}:
    \begin{align}
        & \frac{d}{d\epsilon} p_\epsilon(\lambda_i | pa(\Lambda_i)=\bar{\lambda}_{i-1}) \Big{\rvert}_{\epsilon=0} \nonumber\\ 
        &\quad = p(\lambda_i | pa(\Lambda_i)=\bar{\lambda}_{i-1}) (\mathbb E[S(\Lambda) | \Lambda_i=\lambda_i, pa(\Lambda_i)=\bar{\lambda}_{i-1}] - \mathbb E[S(\Lambda) | pa(\Lambda_i)=\bar{\lambda}_{i-1}]) \label{EIC_key_identity}
    \end{align}
    where $pa(\Lambda_i)$ denotes the parent nodes of variable $\Lambda_i$ and minuscule of a variable denotes the value it takes, we have
    \begin{align}
        \frac{d}{d\epsilon} \Psi(p_\epsilon) \Big{\rvert}_{\epsilon=0} &= \frac{d}{d\epsilon} \Big{\rvert}_{\epsilon=0} \mathbb E_{p_\epsilon} \left[ \mathbb E_{p_\epsilon} \left[ Y' | Z, T'=a \right] \right] \\
        &= \frac{d}{d\epsilon} \Big{\rvert}_{\epsilon=0} \int_{y', z, x} y' \left[ p_\epsilon(y' | z, T'=a) p_\epsilon(z, x) \right] \\
        &= \int_{y', z, x} y' \frac{d}{d\epsilon} \Big{\rvert}_{\epsilon=0} \left[ p_\epsilon(y' | z, T'=a) p_\epsilon(z, x) \right] \quad \text{(dominated convergence)} \\
        &= \int_{y', z, x} y' p(z, x) \frac{d}{d\epsilon} \Big{\rvert}_{\epsilon=0} p_\epsilon(y' | z, T'=a) \\
        &+ \int_{y', z, x} y' p(y' | z, T'=a) \frac{d}{d\epsilon} \Big{\rvert}_{\epsilon=0} p_\epsilon(z, x) \\
        &= \int_w I(t'=a) \frac{p(t' | x)}{p(t' | x)} y' p(y, t | z, x) p(z, x) \frac{d}{d\epsilon} \Big{\rvert}_{\epsilon=0} p_\epsilon(y' | z, t')  \nonumber\\
        &\quad + \int_{y', z, x} y' p(y' | z, T'=a) \frac{d}{d\epsilon} \Big{\rvert}_{\epsilon=0} p_\epsilon(z, x) \\
        &= \int_w \frac{I(t'=a)}{p(t' | x)} y' p(y', t' | z, x) p(y, t | z, x) p(z, x) \left\{ S(w) - \mathbb E\left[ S(W) | y, z, x, t, t' \right]\right\}  \nonumber\\
        &\quad + \int_{y', z, x} y' p(y' | z, T'=a) p(z, x) \left\{ \mathbb E\left[ S(W) | z, x \right] - \mathbb E\left[ S(W) \right]\right\} \\
        &= \int_w \frac{I(t=a)}{p(t | x)} y p(y, t | z, x) p(y', t' | z, x) p(z, x) \left\{ S(w) - \mathbb E\left[ S(W) | y', z, x, t', t \right]\right\}  \nonumber\\
        &\quad + \int_{y', z, x} y' p(y' | z, T'=a) p(z, x) \left\{ \mathbb E\left[ S(W) | z, x \right] - \mathbb E\left[ S(W) \right]\right\} \\
        &= \int_w S(w) \cdot \frac{I(t=a)}{p(t | x)} y p(w) \nonumber\\
        &\quad - \int_w \mathbb E\left[ S(W) | y', z, x, t, t' \right] p(y', z, x, t, t') \cdot \frac{I(t=a)}{p(t | x)} y p(y | z, t) \nonumber\\
        &\quad + \int_{y', z, x} \mathbb E\left[ S(W) | z, x \right] p(z, x) \cdot y' p(y' | z, T'=a) \nonumber\\
        &\quad - \int_{y', z, x} \mathbb E\left[ S(W) \right] \cdot y' p(y' | z, T'=a) p(z, x) \\
        &= \int_w S(w) \left\{\frac{I(t=a)}{p(t | x)} (y - \mathbb E\left[ Y | z, t \right]) + \mathbb E\left[ Y' | z, T'=a \right] - \Psi \right\} p(w)
    \end{align}
    by assumptions of Theorem \ref{variational-ATT} and factorization according to Figure \ref{causal_diagram}. Hence 
    \begin{align}
        \frac{d}{d\epsilon}\Psi(p_\epsilon) \Big{\rvert}_{\epsilon=0} = \left \langle S(W), \frac{I(T = a)}{p(T | X)} (Y - \mathbb E_p\left[ Y | Z, T \right]) + \mathbb E_p\left[ Y' | Z, T'=a \right] - \Psi \right \rangle_{L^2(\Omega; E)}
    \end{align}
    and we have $\tilde{\psi}_p = I(T = a) / p(T | X) \cdot (Y - \mathbb E_p\left[ Y | Z, T \right]) + \mathbb E_p\left[ Y' | Z, T'=a \right] - \Psi$.
\end{proof}

\section{Benchmark Adaptations}
\label{benchmark-adaptation}

\subsection{Autoencoder}
\label{ae-adaptation}
The adapted autoencoder reconstructs the outcome during training similar to a generic autoencoder, but takes treatment and covariates as additional inputs. During test time, we simply plug in the counterfactual treatments along with factual outcomes and covariates to generate the counterfactual outcome predictions.

\subsection{GANITE}
\label{ganite-adaptation}
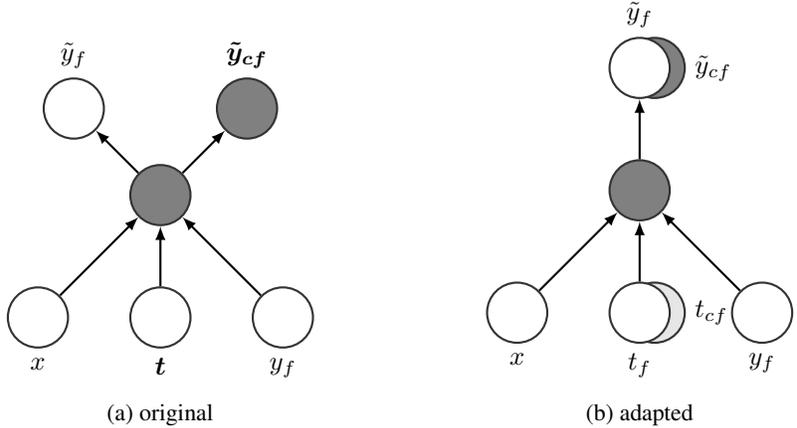
\begin{figure}[hbt!]
    \centering
    \begin{subfigure}[b]{0.45\textwidth}
        \centering
        \begin{tikzpicture}
            \tikzstyle{main}=[circle, minimum size = 8mm, thick, draw =black!80, node distance = 8mm]
            \tikzstyle{connect}=[-latex, thick]
            \tikzstyle{box}=[rectangle, draw=black!100]
              \node[main, fill = white!100] (x) [label=below:$x$] { };
              \node[main, fill = white!100] (t) [right=of x,label=below:$\bm{t}$] { };
              \node[main, fill = white!100] (y) [right=of t,label=below:$y_f$] { };
              \node[main, fill = black!50] (z) [above=of t] { };
              \node[main, fill = white!100] (y_re) [above left=of z,label=above:$\tilde{y}_f$] {};
              \node[main, fill = black!50] (y_cf) [above right=of z,label=above:$\bm{\tilde{y}_{cf}}$] { };
              \path (x) edge [connect] (z)
                    (t) edge [connect] (z)
            		(y) edge [connect] (z)
            		(z) edge [connect] (y_re)
            		(z) edge [connect] (y_cf);
        \end{tikzpicture}
        \caption{original}
        \label{ganite-original}
    \end{subfigure}
    \begin{subfigure}[b]{0.45\textwidth}
        \centering
        \begin{tikzpicture}
            \tikzstyle{main}=[circle, minimum size = 8mm, thick, draw =black!80, node distance = 8mm]
            \tikzstyle{connect}=[-latex, thick]
            \tikzstyle{box}=[rectangle, draw=black!100]
              \node[main, fill = white!100] (x) [label=below:$x$] { };
              \node[main, fill = white!100] (t_f) [right=of x,label=below:$t_f$] { };
              \begin{scope}[on background layer]
                \node[main, fill = black!10] (t_cf) [right=of x,label=right:$t_{cf}$, xshift=2mm] { };
              \end{scope}
              \node[main, fill = white!100] (y) [right=of t_f,label=below:$y_f$] { };
              \node[main, fill = black!50] (z) [above=of t_f] {};
              \node[main, fill = white!100] (y_re) [above=of z,label=above:$\tilde{y}_f$] {};
              \begin{scope}[on background layer]
                \node[main, fill = black!50] (y_cf) [above=of z,label=right:$\tilde{y}_{cf}$, xshift=2mm] { };
              \end{scope}
              \path (x) edge [connect] (z)
                    (t_f) edge [connect] (z)
            		(y) edge [connect] (z)
            		(z) edge [connect] (y_re);
        \end{tikzpicture}
        \caption{adapted}
        \label{ganite-adapted}
    \end{subfigure}
    \caption{GANITE's counterfactual generator. $t_{cf}$ is a random sample of $\bm{t}$, passed into the generator as a part of the input $(x, t_{cf}, y_f)$, separately from input $(x, t_f, y_f)$ of the factual generation.}
    \label{ganite}
\end{figure}

GANITE \cite{yoon2018ganite}'s counterfacutal generator does not scale with a combination of high-dimensional outcome and multi-level treatment, thus here we only input one randomly sampled counterfactual treatment to the generator and correspondingly construct one counterfactual outcome for each sample. See Figure \ref{ganite} for the original and adapted structure of the model.

The discriminator predicts the logits $l_f$, $l_{cf}$ of $y_f$, $\tilde{y}_{cf}$ separately. The cross entropy loss of $(l_f, l_{cf})$ against $(1, 0)$ is then calculated.

\section{Hyperparameter Settings}
\label{hyperparam-setting}

All common hyperparameters of all models are set to the same as the defaults of CPA: an universal number of hidden dimensions $128$; number of layers $6$ (encoder $3$, decoder $3$); discriminator (if any) hidden dimensions $64$ and number of layers $2$; adversarial training (if any) period 3; an universal learning rate $3^{-4}$, decay rate $4^{-7}$ and decay period $45$. For more details regarding the hyperparameter settings of our framework, visit our \href{https://github.com/yulun-rayn/variational-causal-inference}{code repository}.


\end{document}